\pdfoutput=1

\documentclass[11pt]{article}

\usepackage[]{ACL}

\usepackage{times}
\usepackage{latexsym}
\usepackage{adjustbox}
\usepackage[T1]{fontenc}

\usepackage[utf8]{inputenc}

\usepackage{microtype}

\usepackage{inconsolata}

\usepackage{svg}
\usepackage{arydshln}
\usepackage{multirow}
\usepackage{makecell}
\usepackage{lipsum}
\usepackage{caption}
\usepackage{amsmath,amssymb}
\usepackage{hyperref}
\usepackage{enumitem}
\usepackage{bbding}

\usepackage{booktabs}
\newcolumntype{C}[1]{>{\centering\arraybackslash}m{#1}}

\title{SMARTCAL: An Approach to Self-Aware Tool-Use Evaluation and Calibration}

\author{
\textbf{Yuanhao Shen}$^1$, \textbf{Xiaodan Zhu}$^1$, \textbf{Lei Chen}$^2$ \\
$^1$ Department of Electrical and Computer Engineering \& Ingenuity Labs Research Institute \\ Queen's University\\
$^2$ Rakuten Institute of Technology\\
\texttt{\{yuanhao.shen, xiaodan.zhu\}@queensu.ca} \\
\texttt{lei.a.chen@rakuten.com} \\
}

\begin{document}
\maketitle

\begin{abstract}
The tool-use ability of Large Language Models (LLMs) has a profound impact on a wide range of industrial applications. However, LLMs' self-control and calibration capability in appropriately using tools remains understudied. The problem is consequential as it raises potential risks of degraded performance and poses a threat to the trustworthiness of the models. In this paper, we conduct a study on a family of state-of-the-art LLMs on three datasets with two mainstream tool-use frameworks. Our study reveals the tool-abuse behavior of LLMs, a tendency for models to misuse tools with overconfidence. We also find that this is a common issue regardless of model capability. Accordingly, we propose a novel approach, \textit{SMARTCAL}, to mitigate the observed issues, and our results show an average of 8.6 percent increase in the QA performance and a 21.6 percent decrease in Expected Calibration Error (ECE) compared to baseline models.~\footnote{Our code and data are available at 

\href{https://github.com/Henrysyh2000/SMARTCAL}{https://github.com/Henrysyh2000/SMARTCAL}
.
}

\end{abstract}

\section{Introduction}
\label{sec:intro}

The tool-use ability of LLMs has a profound impact on a wide range of applications. Agents that are fine-tuned on various human-computer interaction scenarios such as web browsing \cite{webgpt}, code writing \cite{starcoder}, or even Internet shopping \cite{yang2023autogptonlinedecisionmaking} have been successfully deployed to streamline workflows and boost efficiency in multiple realms within the industry. Recent research has also achieved impressive results by welding various tools into the step-wise reasoning of Retrieval Augmented Generation (RAG), such as a retriever \cite{khattab2022demonstrate}, a database operator \cite{jiang2023structgpt, cheng2022binding, hu2023chatdb}, or a collection of tools \cite{schick2024toolformer, paranjape2023art}. While incorporating tools into LLMs is critical for many applications, \citet{mallen2023not} argue that the tool-use step can negatively impact the performance in some circumstances: e.g., when LLMs have reliable parametric memory. This motivates further studies exploring adaptive retrieval strategies \cite{Asai2023SelfRAGLT, Maekawa2024RetrievalHO}. However, many existing tool-use frameworks rely on either passive in-context learning from existing few-shot examples \cite{paranjape2023art, khattab2022demonstrate, hu2023chatdb} or fine-tuning on dedicated datasets \cite{toolken, schick2024toolformer, jiang2023structgpt, cheng2022binding}. The absence of a model's active thinking in tool-use thus leaves a crucial question under-studied: \textit{Are LLMs aware of \textbf{when} to use \textbf{which} tool?}




To understand the performance of using tools, we conduct a series of experiments under the scenario of open domain QA \cite{roberts-etal-2020-much}. Our results raise  concerns related to the above question: the tracking of LLM tool usage across \texttt{ChatGPT} series \cite{openai_gpt} and \texttt{llama-3-instruct} on Entity Questions data \cite{sciavolino-etal-2021-simple} shows that on average, a model misuses one or more types of tools in over 20\% of its total reasoning steps. Additionally, when the model is asked to report its confidence in selecting a certain tool within each step, more than 90\% of its stated confidence falls in the confidence bin where the reported confidence level is higher than the actual answering accuracy, indicating the model's overconfidence with respect to tool choice. The bottom part of the first two columns in Figure \ref{fig:framework} demonstrates such tool-abuse phenomenon. 

In this paper, we propose \textit{SMARTCAL}, a novel approach to helping mitigate tool-abuse. \textit{SMARTCAL} consists of three components \textit{(i) Self-Evaluation} (SE), \textit{(ii) Confidence Prior Collection} (CPC), and \textit{(iii) Augmented Reasoning} (AR), which mitigate tool-misuse and provide a more reliable calibration performance. Deployment of \textit{SMARTCAL} on two different tool-use frameworks, ART~\cite{paranjape2023art} and DSP~\cite{khattab2022demonstrate},  shows that it is able to derive an efficient strategy on tool-use and provides better calibrated answers.


To the best of our knowledge, this is among the first efforts focused on investigating the calibration of LLM-based tool-use. Fostering proper use of tools is considered to be important for many applications that emphasize the alignment of LLMs \cite{shen2024towards}. Our contributions are summarized as follows: We observe tool-abuse in LLMs, which includes tool-misuse behavior and an inaccurate evaluation of verbalized confidence scores. We show that degradation in tool-use calibration remains a common issue regardless of increasing model capabilities. We introduce \textit{SMARTCAL}, a novel framework that aims to mitigate tool abuse. \textit{SMARTCAL} achieves an average of 8.6 percent increase in the QA performance and a 21.6 percent decrease in Expected Calibration Error (ECE) compared to baseline models. 



\section{\textit{SMARTCAL}: A Tool-Use Recalibration Approach}
\label{sec:framework}

\begin{figure*}[!h]
    \centering
    \includegraphics[width=\linewidth]{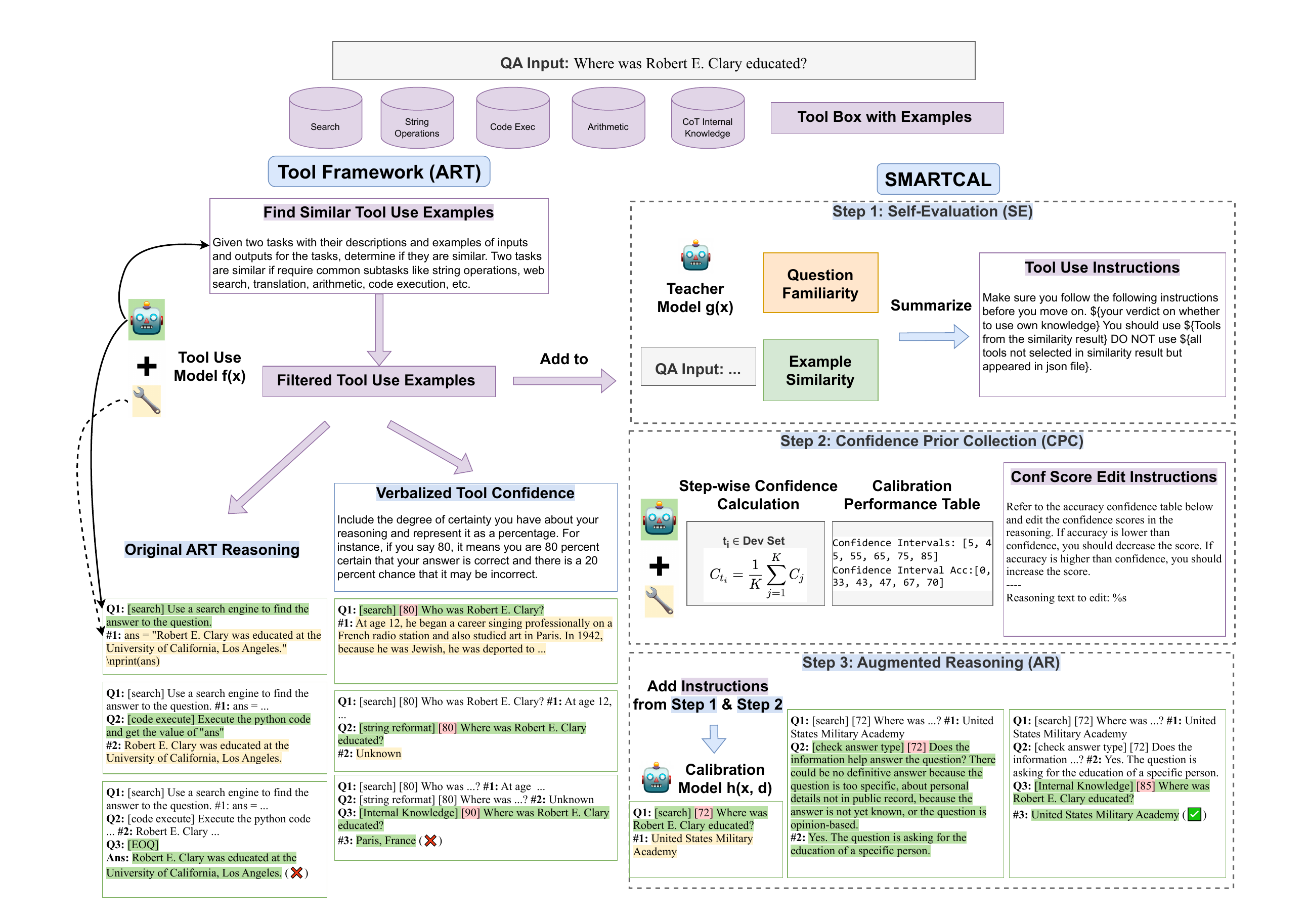}
    \caption{Comparison between ART \cite{paranjape2023art}, ART (V), and \textit{SMARTCAL} on the complex QA task. ART (V) introduces verbalized confidence elicitation. \textit{SMARTCAL} includes three steps to mitigate tool-abuse.}
    \label{fig:framework}
\end{figure*}


Motivated by the self-verification feature that constitutes the reasoning capability in a multi-agent system \cite{pezeshkpour2024reasoning}, we introduce a novel framework \textit{SMARTCAL} that helps control tool-misuse based on multiple LLM agents. Different from existing approaches that emphasize in-context learning from demonstrations such as Automatic Multi-step Reasoning and Tool-use (ART) \cite{paranjape2023art} shown in the left column in Figure \ref{fig:framework} and Demonstrate Search Predict (DSP) \cite{khattab2022demonstrate}, \textit{SMARTCAL} incorporates extra evaluation steps to examine the legitimacy of tool usage within each step. Additionally, compared to existing tool-use frameworks where each step is controlled by a single agent, \textit{SMARTCAL} features an enhanced pipeline that promotes the collaboration among the agents, ensuring accurate and reliable tool usage during step-wise reasoning. Specifically, when prompted with an input task, \textit{SMARTCAL} first derives an optimized strategy about \textit{when} to use \textit{which} tool. Then, the collaboration between specialized agents actively interferes with and corrects potential tool-abuse risks in the enhanced pipeline. Table \ref{tab:capability} shows a comparison of \textit{SMARTCAL} with DSP and ART.

\begin{table}[h]
    \centering
    \renewcommand{\arraystretch}{1.3} 
    \small
    \begin{tabular}{C{3cm}|C{1cm}|C{1cm}|C{1.5cm}}
          \toprule
          \toprule
          \textbf{Capability} & \textbf{DSP} & \textbf{ART} & \textbf{SMARTCAL (Ours)} \\
          \midrule
          Retrieval Augmented Generation & \checkmark & \checkmark  & \checkmark \\
          Use Multiple Tools & & \checkmark  & \checkmark \\
          Report Tool Confidence & & & \checkmark \\
          Tool Confidence Calibration & & & \checkmark \\
          Tool selection evaluation & & & \checkmark \\

         \bottomrule
         \bottomrule
    \end{tabular}
    \caption{Comparing \textit{SMARTCAL} with existing frameworks that is capable of using tools in reasoning.}
\label{tab:capability}
\end{table}

We provide an overview of our framework in Figure \ref{fig:framework}, which depicts ART as an example of tool-use frameworks. Meanwhile, \textit{SMARTCAL} is also compatible with existing tool-use frameworks that incorporate in-context learning with few-shot examples. In our experiments, we report \textit{SMARTCAL} results on both ART and DSP. We also derive ART (V) and DSP (V) that incorporate verbalized calibration and compare the accuracy with \textit{SMARTCAL}. Specifically, \textit{SMARTCAL} has three components: \textit{(i) Self-Evaluation} (SE) provides tool-use instructions, \textit{(ii) Confidence Prior Collection} (CPC) collects model-specific confidence prior, and \textit{(iii) Augmented Reasoning} (AR) combines the previous results into a collaborative pipeline. These components aim to mitigate tool-abuse from the following perspectives: (1) introducing constraints on tool usage from self-evaluation and (2) incorporating tool confidence prior into the reasoning process. 


\subsection{Self-Evaluation (SE)}
The SE component employs a teacher model $g(x)$ to conduct self-evaluation, where we denote $x$ as the input task plus few-shot tool-use examples. Taking as an example the question ``Where was Robert E. Clary educated?'', \textit{SMARTCAL} applies $g(x)$ based on two dimensions: (1) $g_{fam}(x)$ for Task Familiarity and (2) $g_{sim}(x)$ for Example Similarity. Familiarity evaluation focuses on assessing whether the parametric memory itself is already sufficient to handle the task. If the task is solvable using model's own knowledge, $g_{fam}(x)$ will include ``\texttt{[Internal Knowledge]}'' as an option and tell the model to be more careful when using tools. Otherwise, $g_{fam}(x)$ will provide a verdict to encourage tool-use model to use tools. For similarity evaluation, it focuses on extracting specific tools used in the selected examples and picks out the ones that are useful to solve the task. In this example, $g_{sim}(x)$ extracts ``\texttt{[search]}'' and ``\texttt{[check answer type]}'' as the useful tools from the filtered tool-use examples in the similarity evaluation, and the familiarity evaluation results encourage the tool-use model $f(x)$ to incorporate ``\texttt{[Internal Knowledge]}'' as an option to answer the question based on the tool-use context. Both familiarity and similarity results are then summarized into an aggregated instruction $I$ that the model can follow to handle the task. Detailed prompts can be found in Appendix \ref{sec:appendix_prompts}.



\subsection{Confidence Prior Collection (CPC)}

Building on the SE, the CPC component collects model-specific prior calibration information in order to provide more accurate tool confidence scores. We pre-run a heldout subset $D$ with tool-use model $f(x)$, and add self-evaluation instructions $I$ in the reasoning process. Motivated by recent studies that achieve decent calibration performance through verbalized confidence elicitation \cite{lin2022teaching, xiong2023llms, tian2023just}, we adapt this technique into step-wise confidence elicitation during the tool-use phase of the agent. Denote a dev set task $t_i \in D$ with $K$ steps of tool-use, each step containing verbalized confidence $C_j$. We calculate the average $C_{t_i}$ to represent the agent's overall confidence in using tools. The answers from $D$ with calculated confidence scores are binned at a preset stepsize and the accuracy is calculated respectively. The calibration results are then organized as a confidence-accuracy lookup table $\{conf\_level, acc\}$. The formula of confidence calculation and confidence prior structure are shown in the CPC block of Figure \ref{fig:framework}. 

The performance of the heldout dataset is regarded as the approximation of the underlying confidence-accuracy distribution on the test dataset. The results will serve as the prior reference for the model when editing the output tool confidence using a calibration model.



\subsection{Augmented Reasoning (AR)}

Once we obtain the self-evaluation results and confidence prior, the AR component will integrate the previous results in the following procedure. First, self-evaluation instruction $I$ is generated by the teacher model $g(x)$ and is augmented on selected tool-use examples. Then, the tool-use model $f(x)$ is called to output the intermediate reasoning contexts with controlled usage of tools and verbalized tool confidence. Finally, the confidence prior $D$ expressed in a lookup table is used to detect and correct overconfidence or underconfidence on tool usage. We describe the reasoning pipeline of AR using the QA example in Figure \ref{fig:framework}: tool-use agent outputs reasoning context with more controlled tool usage following instructions in the SE module to include ``\texttt{[search]}'' and ``\texttt{[Internal Knowledge]}''. Calibration model $h(x, d)$ interacts with both tool-use agent result and confidence prior to provide edited confidence evaluations and the final answer to the question.

\section{Experiment Setup}

\begin{figure}[t]
    \centering
    \includegraphics[width=\linewidth]{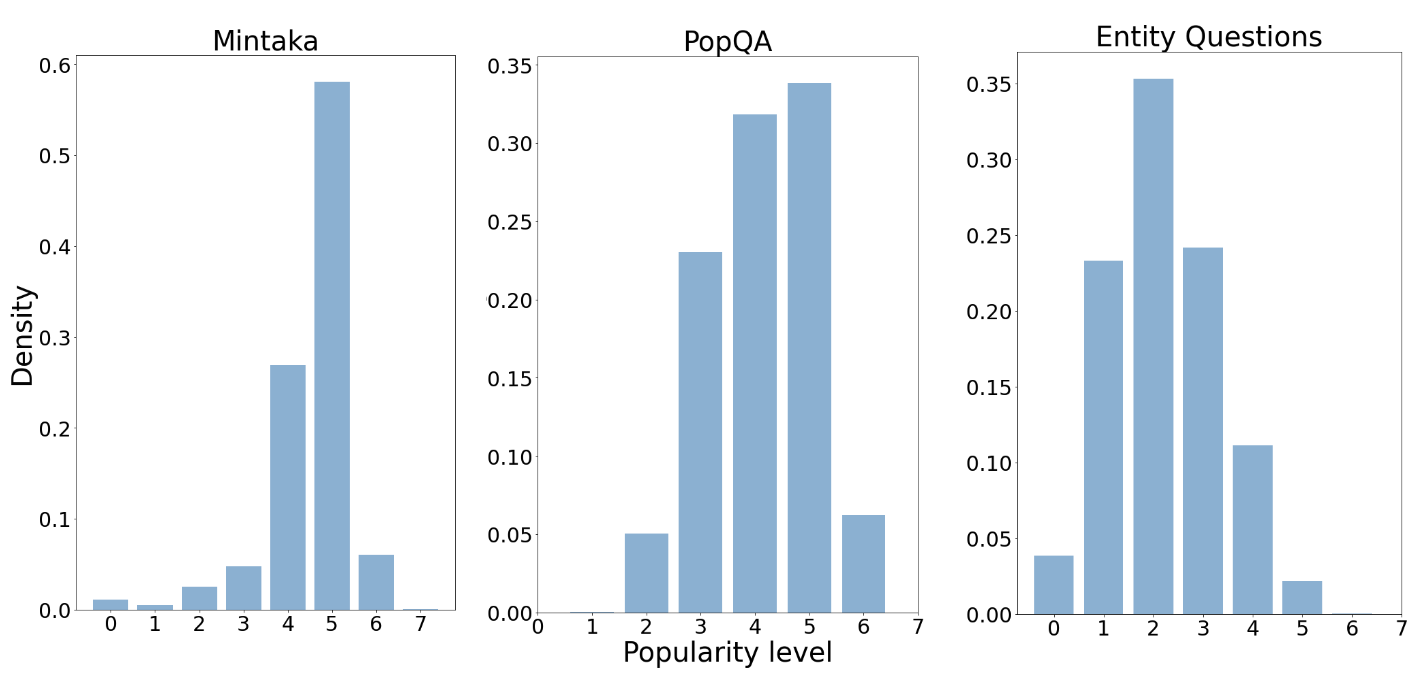}
    \caption{Distribution of entity popularity for Mintaka, PopQA, and Entity Questions dataset.}
    \label{fig:data}
\end{figure}

\paragraph{Tasks and Datasets.}
We perform our experiments under the open-domain QA setup \cite{roberts-etal-2020-much} using three benchmark datasets:  Mintaka \cite{sen2022mintaka}, PopQA \cite{mallen2023not}, and Entity Questions \cite{sciavolino-etal-2021-simple}. A histogram of the popularity distribution of these datasets can be found in Figure \ref{fig:data}.

Following the findings from \citet{mallen2023not} which point out that retrieval is mandatory when the model lacks parametric memory, we sample the tail distribution of the three datasets in Figure \ref{fig:data} to simulate the setting when tool-use agents are dealing with out-of-scope knowledge. Specifically, we set dedicated threshold based on each dataset to construct the low popularity subset.Appendix \ref{sec:appendix_dataset} offers a detailed description as well as the augmentation of popularity information of the three datasets. 

\begin{itemize}
    \item \textbf{Mintaka.} \citet{sen2022mintaka} collect a human elicited dataset that contains QA pairs that span eight categories. This dataset has received notable attention in recent studies \cite{li2024enhanced, sun2024pyramid} to provide benchmark and insight in a real-world setting about how models behave when choosing tools to augment their reasoning.

    \item \textbf{PopQA \& Entity Questions.} PopQA \cite{mallen2023not} and Entity Questions \cite{sciavolino-etal-2021-simple} are two synthetic datasets that contain knowledge intensive QA tasks. The questions are organized in a triplet containing subject, relationship, and object, which are wrapped in a fixed QA template.
    
\end{itemize}


\paragraph{Models.} Experiments are run on two \texttt{ChatGPT} models and \texttt{llama-3-70b-instruct}. We select the more advanced \texttt{gpt-4-turbo} as the teacher model and \texttt{gpt-3.5-instruct-0914} for better instruction following ability as the calibration model in \textit{SMARTCAL} framework. Appendix \ref{sec:appendix_models} includes model details in our experiments. 


\paragraph{Evaluation Metrics.}

For the QA performance, we report the Exact Match (EM) score, and for calibration metric, we use Expected Calibration Error (ECE) \cite{naeini2015obtaining, Obadinma2021Class}. Details can be found in Appendix \ref{sec:appendix_metrics}.

\section{Experiment Results and Analysis}
\label{sec:res_and_ana}

\begin{table*}[h]
    \renewcommand{\arraystretch}{1} 
    \centering
    \begin{tabular}{c|C{4cm}|C{1cm}C{1.4cm}|C{1.8cm}|C{1cm}C{1.4cm}|C{1.8cm}}
         \toprule
         \multicolumn{1}{c}{} & \textbf{Models} & \multicolumn{1}{c}{\textbf{DSP}} & \textbf{DSP (V)} & \textbf{DSP+ SMARTCAL} & \multicolumn{1}{c}{\textbf{ART}} & \textbf{ART (V)} & \textbf{ART+ SMARTCAL} \\
         \midrule
         \multirow{3}{*}{\makecell[c]{\rotatebox[origin=c]{90}{Mintaka}}} 
         & \texttt{gpt-3.5-turbo} & 0.417 & 0.464 & \textbf{0.490} & 0.497 & 0.477 & \textbf{0.517} \\
         & \texttt{gpt-4} & 0.371 & 0.358 & \textbf{0.450} & 0.470 & 0.550 & \textbf{0.596} \\
         & \texttt{llama-3-70b-instruct} & 0.377 & 0.464 & \textbf{0.464} & 0.623 & 0.603 & \textbf{0.629} \\
        \midrule
         
         \multirow{3}{*}{\rotatebox[origin=c]{90}{PopQA}} 
         & \texttt{gpt-3.5-turbo} & 0.417 & 0.401 & \textbf{0.591} & 0.016 & 0.064 & \textbf{0.131} \\
         & \texttt{gpt-4} & 0.374 & 0.371 & \textbf{0.613} & 0.553 & 0.552 & \textbf{0.557} \\
         & \texttt{llama-3-70b-instruct} & 0.361 & 0.360 & \textbf{0.362} & 0.529 & 0.518 & \textbf{0.533} \\
         \midrule
         
         \multirow{3}{*}{\rotatebox[origin=c]{90}{Entity Q.}} 
         & \texttt{gpt-3.5-turbo} & 0.503 & 0.481 & \textbf{0.574} & 0.423 & 0.557 & \textbf{0.570} \\
         & \texttt{gpt-4} & 0.506 & 0.505 & \textbf{0.603} & 0.448 & 0.449 & \textbf{0.635} \\
         & \texttt{llama-3-70b-instruct} & 0.445 & 0.490 & \textbf{0.490} & 0.526 & 0.574 & \textbf{0.606} \\

         \bottomrule

    \end{tabular}
    \caption{QA accuracy comparison of \textit{SMARTCAL} implementation on three datasets using two frameworks. \texttt{gpt-3.5-turbo} and \texttt{gpt-4} results are accessed between Feburary 2024 to June 2024.}
    \label{tab:acc_res}

\end{table*}

\begin{table*}[!h]
    \centering
    \renewcommand{\arraystretch}{1} 
    \setlength{\tabcolsep}{3pt}
    \begin{tabular}{c|cc|cc|cc}
          \toprule
           & \multicolumn{2}{c}{Mintaka} & \multicolumn{2}{c}{PopQA} & \multicolumn{2}{c}{Entity Ques} \\
          \midrule
          Models & ART (V) & SMARTCAL & ART (V) & SMARTCAL & ART (V) & SMARTCAL \\
          \midrule
         \texttt{gpt-3.5-turbo} & 0.451 & \textbf{0.445}  & \textbf{0.010} & 0.087  & 0.513 & \textbf{0.507} \\
         \texttt{gpt-4} & 0.263 & \textbf{0.169} & 0.261 & \textbf{0.201}  & 0.236 & \textbf{0.096}  \\
         \texttt{llama-3-70b-instruct} & 0.335 & \textbf{0.145} & 0.172 & \textbf{0.113} & 0.133 & \textbf{0.103}  \\

         \bottomrule
    \end{tabular}
    \caption{Calibration performance (ECE) of ART plus \textit{SMARTCAL} on three datasets. Note that for ECE scores, the lower the better. \texttt{gpt-3.5-turbo} and \texttt{gpt-4} results are accessed between Feburary 2024 to June 2024.}
    \label{tab:ece_res}
\end{table*}

\subsection{Overall QA Performance}
We conduct our study on two tool-use frameworks, DSP \cite{khattab2022demonstrate} and ART \cite{paranjape2023art}. In addition to the original setting, we also introduce verbalized confidence elicitation settings of the two frameworks denoted as ART (V) and DSP (V). In Table \ref{tab:acc_res}, we report both settings and compare them in conjunction with \textit{SMARTCAL}. We can see that when \textit{SMARTCAL} is augmented on both frameworks, it either surpasses or performs on par in terms of QA performance compared to the baseline setting as well as the verbalized calibration setting. The baseline settings of DSP achieves an average of 41.9\% on all datasets, while ART has an average accuracy of 45.4\%. In comparison, \textit{SMARTCAL} achieves 51.5\% when adapted to DSP and 53.0\% when adapted to ART, with an average advantage of 8.6\% in accuracy improvement. We also observe an excessive inferiority in QA accuracy for \texttt{gpt-3.5-turbo} on PopQA dataset, where the model is unwilling to answer most questions. We elaborate this observation in Appendix \ref{sec:appendix_qa_analysis}.


\subsection{Calibration Performance}
Table \ref{tab:ece_res} presents the ECE score with ART (V) and \textit{SMARTCAL}. For almost all experiments, ART (V) yields a higher calibration error, with an average ECE of 0.264. \textit{SMARTCAL} achieves an average ECE of 0.207 on the testing datasets, with an average of 21.6\% fewer errors in the confidence alignment. Again for \texttt{gpt-3.5-turbo}, we observe inferiority in ART (V) when tested on PopQA data. We elaborate this observation in Appendix \ref{sec:appendix_cali_analysis}. 

\begin{figure}[!h]
    \centering
    \includegraphics[width=\linewidth]{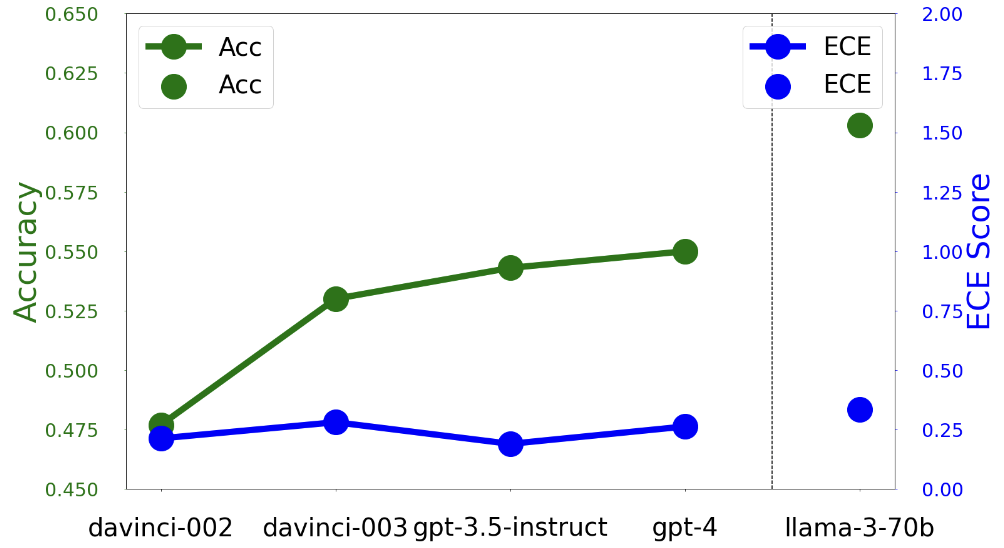}
    \caption{ECE and QA accuracy trend comparison on Mintaka dataset. ECE scores remain stable despite increasing model capability. }
    \label{fig:ece_acc_comp}
\end{figure}

In addition to the ECE performance in Table \ref{tab:ece_res}, we also record QA accuracy and ECE performance on less capable GPT models and create a trend plot on Mintaka data in Figure \ref{fig:ece_acc_comp}. Interestingly, we find qualitatively from the plot that ECE results remain stable with fluctuations between 0.15 to 0.50, despite increasing model capability. In contrast, QA accuracy continues to improve from 47\% to near 60\% with an evolving model ability.


\subsection{Detailed Analysis}

Are LLMs aware of \textit{when} to use \textit{which} tool? Our results above raise concerns that tool-misuse poses a threat to the QA performance. Also, despite a certain level of awareness, LLMs lack more targeted tool-use calibration methods. Thus, \textit{SMARTCAL} aims to provide a preliminary solution from the two perspectives as detailed below. 

\begin{figure}[h]
    \centering
    \includegraphics[width=\linewidth]{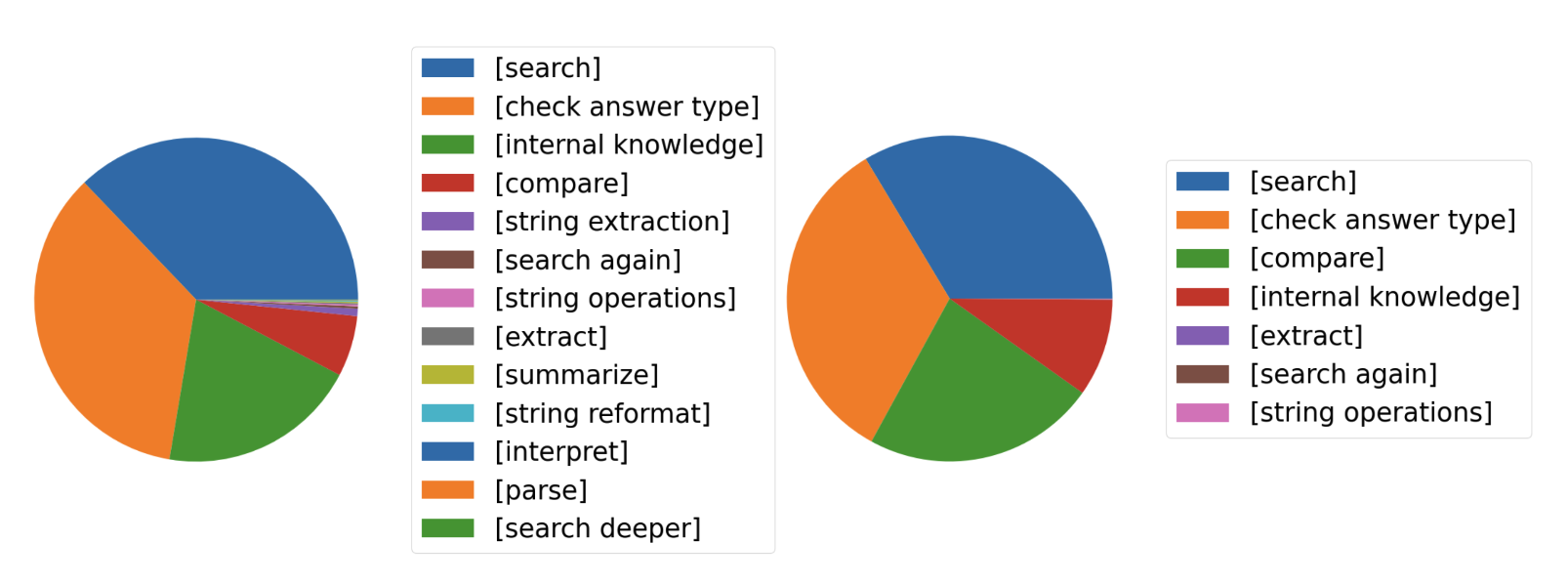}
    \caption{Tracking of tool usage from Entity Questions data \cite{sciavolino-etal-2021-simple} on \texttt{GPT-4}. Diagram on the left is the original ART tool usage, on the right is \textit{SMARTCAL} tool usage breakdown.}
    \label{fig:tool_abuse}
\end{figure}

\paragraph{\textit{SMARTCAL} improves performance by mitigating tool-misuse.} Previous work has shown the necessity of retrieval under low popularity context \cite{mallen2023not}. We further show that tool-misuse may also exert a negative effect on the answering accuracy. Figure \ref{fig:tool_abuse} shows a comparison of \texttt{gpt-4} between ART and \textit{SMARTCAL} on how tools are used in the Entity Questions data. A full comparison of all datasets is included in the tool usage collection section in Appendix \ref{sec:appendix_tool_collection}. We can see that ART tends to use a variety of tools, many of which are not providing useful contexts, resulting in a QA accuracy of 0.448. On the other hand, \textit{SMARTCAL} reduces the use of unnecessary tools significantly via the SE step, increasing the accuracy to 0.635. Thus, the introduction of those excessive tools, if not properly used with the corresponding levels of confidence, could negatively influence the QA accuracy.



\begin{figure}[h]
    \centering
    \includegraphics[width=\linewidth]{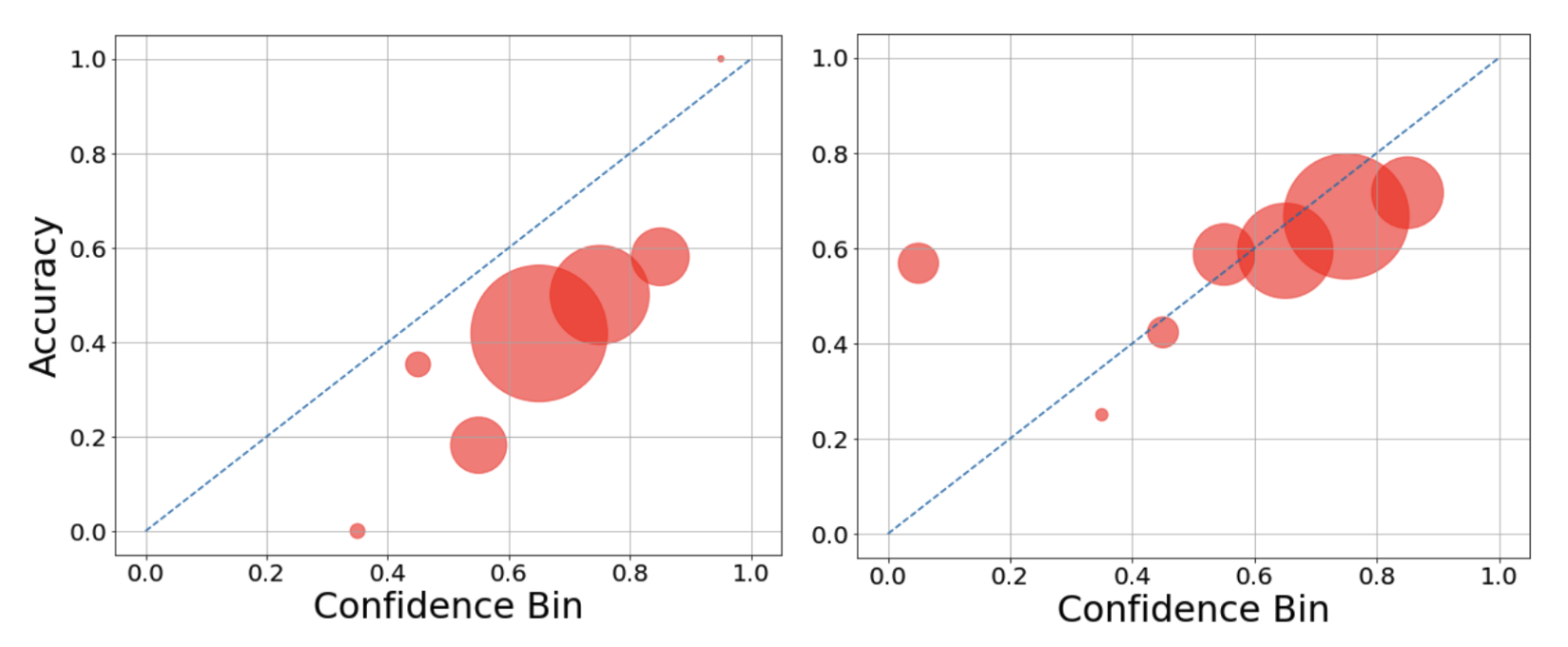}
    \caption{Calibration performance comparison of \texttt{GPT-4} on Entity Questions data with ART (V) on the left and \textit{SMARTCAL} on the right.}
    \label{fig:ece_comp}
\end{figure}

\paragraph{\textit{SMARTCAL} recalibrates tool usage confidence via agent collaboration.} The augmented reasoning step in \textit{SMARTCAL} takes advantage of the calibration results from the heldout dataset. By using the results as a prior, the calibration agent in \textit{SMARTCAL} is able to interact with contexts generated by the tool-use agent and to edit the confidence score stated in the verbalized approach, thereafter providing more reliable tool-use confidence scores. Figure \ref{fig:ece_comp} provides a comparison of the reliance plot of \texttt{gpt-4} on Entity Questions data. Note that the zero confidence interval represents the questions where regular expressions failed to extract a valid confidence score from the agent’s reasoning history. A full comparison of calibration performance plot can be found in Appendix \ref{sec:appendix_curve}


\subsection{Ablation Study}
In this section, we further study the relative importance of each component within \textit{SMARTCAL}. We choose the ART setup to conduct ablations using the Mintaka data on three models. Specifically, we  mask either the SE or the CPC component in \textit{SMARTCAL} and measure the QA accuracy and ECE respectively. Table \ref{tab:acc_abla} and Table \ref{tab:ece_abla} showcase the results.

In terms of the SE module, we find it useful both in increasing QA performance as well as in lowering calibration error. Among the three models tested when CPC is masked, SE module achieves an average of 2.9\% increase in QA accuracy compared with the baseline when both SE and CPC are disabled. It can also be observed that adding self-evaluation also helps the model to be more aware of tool-use confidence reports. The second column in Table \ref{tab:ece_abla} with SE module enabled achieves an average of 21.6\% lower in calibration error compared to the baseline. 

For the CPC module, it can be seen from the ablation results that it further helps lower the calibration error, with an average of 39.4\% lower in calibration error when comparing with the baseline in Table \ref{tab:ece_abla}. This further suggests that with the integration of confidence prior, it helps the model become more informed on providing reliable confidence scores. 

\begin{table*}[!h]
    \centering
    \renewcommand{\arraystretch}{1} 
    \begin{tabular}{c|c|c|c}
          \toprule
            \textbf{Models} & \textbf{w/o CPC, w/o SE} & \textbf{w/ CPC, w/o SE} & \textbf{w/o CPC, w/ SE} \\
          \midrule
         \texttt{gpt-3.5-turbo-0613} & 0.536 & 0.536  & \textbf{0.576} \\
         \texttt{gpt-4} & 0.576 & 0.576 & \textbf{0.589} \\
         \texttt{llama-3-70b-instruct} & 0.623 & 0.623 & \textbf{0.656} \\
         \bottomrule
    \end{tabular}
    \caption{QA accuracy of SE and CPC components in SMARTCAL using Mintaka data.}
    \label{tab:acc_abla}
\end{table*}

\begin{table*}[!h]
    \centering
    \renewcommand{\arraystretch}{1} 
    \begin{tabular}{c|c|c|c}
          \toprule
            \textbf{Models} & \textbf{w/o CPC, w/o SE} & \textbf{w/o CPC, w/ SE} & \textbf{w/ CPC, w/o SE} \\
          \midrule
         \texttt{gpt-3.5-turbo-0613} & 0.233 & 0.161  & \textbf{0.096} \\
         \texttt{gpt-4} & 0.245 & 0.244 & \textbf{0.126} \\
         \texttt{llama-3-70b-instruct} & 0.110 & \textbf{0.073} & 0.098 \\
         \bottomrule
    \end{tabular}
    \caption{Calibration performance (ECE) of SE and CPC components in SMARTCAL using Mintaka data.}    
    \label{tab:ece_abla}
\end{table*}



\section{Related Work}

\paragraph{Retrieval Augmented Generation (RAG).} 
Task decomposition techniques \cite{wei2022chain, yang2022re3, ozturkler2022thinksum, kazemi2022lambada, reppert2023iterated, creswell2022selection, puerto2024codepromptingelicitsconditional, fang-etal-2024-dara} augmented with retrieved contexts in knowledge-intensive NLP tasks \cite{karpukhin2020dense, webgpt, li-etal-2023-finance} have been shown to be very effective in various complex NLP tasks. Recent work \cite{jiang2023structgpt, cheng2022binding, hu2023chatdb} has augmented Chain-of-Thought with external database operations to facilitate LLM reasoning on tabular data. Knowledge distillation approaches~\cite{schick2024toolformer, paranjape2023art, cai2023large} have also been proposed to teach LLM to create and use tools in order to enhance reasoning performance. 

\paragraph{Selective Retrieval Methods in RAG.} 
Recent work empirically reveals that RAG has a negative impact on QA performance when LLMs have better memorization of popular factual knowledge \cite{mallen2023not}. This work further motivates an exploration into selective retrieval methods, including fine-tuning smaller models to provide factuality checking and ranking \cite{Tian2023FinetuningLM} and generating retrieval evaluations to avoid excessive and noisy contexts \cite{Asai2023SelfRAGLT, Maekawa2024RetrievalHO}, paving the way for more versatile and efficient RAG strategies.


\paragraph{Calibration in LLMs.} Recent attempts to study LLM calibration often include adversarial attacks \cite{obadinma2024calibrationattackscomprehensivestudy}, while other approaches have connected this notion with confidence-level elicitation \cite{guo2017calibration, minderer2021revisiting, xiong2023llms}. Current approaches include verbalized confidence elicitation \cite{lin2022teaching}, which asks for a confidence score directly when answering a factual question. \citet{xiong2023llms} take a step further by combining this verbalized approach with self-consistency and propose a hybrid confidence elicitation framework. However, existing work focuses more on single-step reasoning calibration on factual information, overlooking its efficacy under the multi-step context of using tools.

\section{Conclusion}
In this paper, we identify tool-abuse in LLM reasoning, which involves a combination of tool-misuse and degraded tool calibration performance. We also observe a consistently high calibration error regardless of increasing model scales. We then propose a novel framework \textit{SMARTCAL} to mitigate this issue. To our knowledge, this is among the first efforts to study the topic of recalibration for LLM-based tool-use.

\section{Limitations}
As for our future work, we would like to extend the proposed method to complex multi-step reasoning tasks. Also, our experiments and results are limited to a subset of the existing datasets to observe tool-misuse behavior. It would be interesting to observe if such behavior remains consistent in more complex datasets elicited by humans that contain multiple reasoning paths.


\bibliography{custom}
\bibstyle{acl_natbib}

\appendix
\section{Appendix}

\subsection{Dataset Details}
\label{sec:appendix_dataset}

\subsubsection{Mintaka} \citet{sen2022mintaka} collect a human elicited dataset that requires complex reasoning with an amalgamation of eight distinct symbolic operations, spanning more than eight different topics, totaling a number of 20,000 labeled questions. We augment the Mintaka dataset with popularity information \footnote{We use log-based weekly pageviews from Wikidata API to obtain the popularity level from the questions. We define the entities with log pageviews less than two as low popularity, and higher than four as high popularity.} and make a test set that contains 151 questions with low popularity. Since the number of questions in the training set with low popularity is more limited (50 questions), we randomly sample 200 data to construct the dev set in confidence calibration.

\subsubsection{PopQA \& Entity Questions} PopQA \cite{mallen2023not} and Entity Questions \cite{sciavolino-etal-2021-simple} are two synthetic datasets that contain knowledge intensive QA tasks. The entities are organized in a triplet containing subject, relationship, and object wrapped in a fixed template to form a question. Given that the two datasets all contain the Wikidata-scraped popularity information, we directly filter out the low popularity section within those datasets, providing a total of 2,349 questions in test set. For the dev set in confidence calibration, we sample 200 questions from PopQA and 500 questions from Entity Questions that are of low popularity in the training set.

\subsection{Experiment Details}
\label{sec:appendix_experiment}

\subsubsection{Models}
\label{sec:appendix_models}

\paragraph{InstructGPT.} First released in November 2022, \texttt{InstructGPT} is a series of models that is trained by OpenAI to conduct text completion tasks. The original \texttt{text-davinci} series is considered less capable at understanding instructions . OpenAI deprecated their older \texttt{text-davinci} series and updated their instruct models in September 2023 with \texttt{gpt-3.5-turbo-instruct}, making it more capable at following instructions. 

\paragraph{ChatGPT.} We also include a spectrum of models with different capabilities in the \texttt{ChatGPT} series \cite{openai_gpt}, including \texttt{gpt-3.5-turbo}, \texttt{gpt-4}, and \texttt{gpt-4-turbo}. 

\paragraph{Llama-3 Instruct.} As an updated version from \texttt{llama-2} \cite{touvron2023llama}, \texttt{llama-3} is trained with more recent corpora from various sources and achieves a better performance in various benchmarks. Different from the GPT family, Llama models are completely open-source. \texttt{llama-3-instruct} features two models divided by parameter sizes, including \texttt{llama-3-instruct-8b} and \texttt{llama-3-instruct-70b}.

In \textit{SMARTCAL}, \texttt{gpt-3.5-instruct-0914} is used for similar task selection in the ART framework. For the teacher model in the SE module described in section \ref{sec:framework}, we select \texttt{gpt-4-turbo} to provide self-evaluation results. For the calibration model in the AR module, we employ \texttt{gpt-3.5-instruct-0914} for better instruction following to edit the tool-use context. The temperature of all models tested is set to 0.7 in both ART and DSP modules according to the best reported results from \citet{paranjape2023art} and \citet{khattab2022demonstrate}. The max token length for each reasoning step in ART is set to be 500 and it is 800 in DSP. For maximum steps within the reasoning process, ART has a maximum of 10 steps, while DSP is set to 3 steps.

\subsubsection{Evaluation Metrics}
\label{sec:appendix_metrics}
In our experiments, we use a more generic version of Exact Match (EM). Denote the answer from the model as $a_M$, and the label as $L$. The answer is considered correct if:
\begin{equation}
    a_M \subseteq L \cup L \subseteq a_M
\end{equation}

For calibration evaluation, we use the ECE score. ECE essentially describes the deviation between the model's stated confidence and its true accuracy. It bins the answers according to the model's stated confidence and calculates the average first norm distance between the QA accuracy within the bin and the confidence score. Denoting $a_M$ as the answer from the model, and $p_M$ as the probability assigned by the model that $a_M$ is correct, $p$ is the actual QA accuracy in this confidence bin. ECE is calculated as follows: 
\begin{equation}
    \mathbb{E}_{p_M}[|\mathbb{P}(a_M| p_M=p) - p|]
\end{equation}

\subsection{Result Analysis}
\label{sec:appendix_analysis}

\subsubsection{Tool-Use Behavior Analysis}
\label{sec:appendix_qa_analysis}

\begin{figure}[!h]
    \centering
    \includegraphics[width=\linewidth]{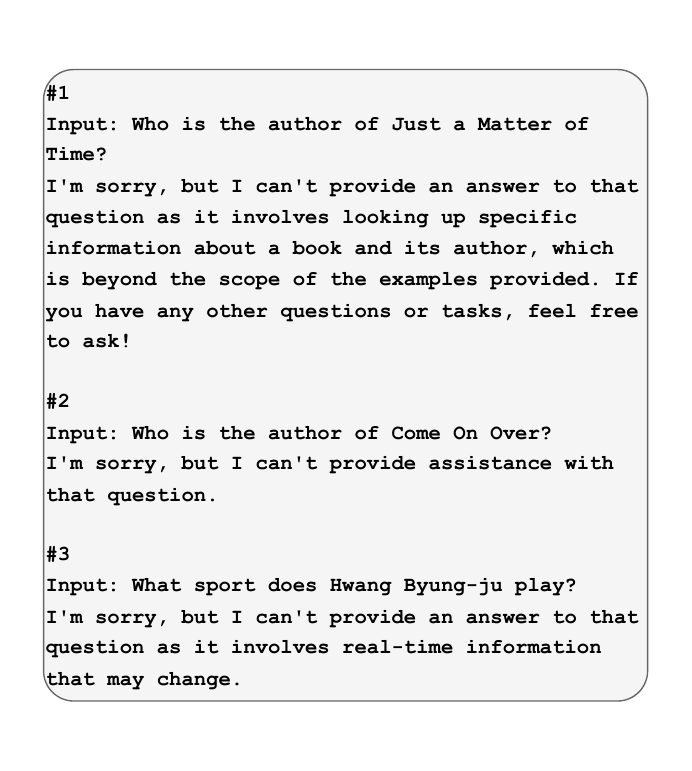}
    \caption{Examples of \texttt{gpt-3.5-turbo} reasoning history on PopQA dataset using ART.}
    \label{fig:error_sample}
\end{figure}

In this section, we provide more detailed analysis following the reported results in Section \ref{sec:res_and_ana}. As we mentioned earlier, \texttt{gpt-3.5-turbo} achieves unexpectedly low QA accuracy on PopQA dataset on ART framework. We provide several examples that record the history of \texttt{gpt-3.5-turbo} reasoning when tested on PopQA data in Figure \ref{fig:error_sample}. We can see from the history that for most of the tested questions, \texttt{gpt-3.5-turbo} refuses to provide a concrete answer that follows the few-shot structure in the ART framework. Instead, it either states that the question needs extra information or it simply can't assist in answering the question. Based on our results, this answer pattern is common regardless of other settings, including the incorporation of verbalized confidence elicitation and \textit{SMARTCAL}.

\subsubsection{Tool-Use Calibration}
\label{sec:appendix_cali_analysis}
Following the calibration performance in Table \ref{tab:ece_res}, we observe that under the schema of the verbalized confidence elicitation, the model tends to assign a fixed and consistent confidence score (i.e. 80\% confidence whenever it uses the \texttt{[search]} tool in the reasoning step), which in turn makes the aggregated tool-use confidence clustering around a certain confidence interval. This observation is consistent with the results obtained by \citet{lin2022teaching}. Additionally, the unexpected behavior elaborated in Appendix \ref{sec:appendix_qa_analysis} also affects the calculation of calibration performance. When calculating average tool confidence, we default the confidence score to zero when we fail to extract tool usage from the generated reasoning history. An edge case of such a setting is when the overall QA accuracy is also extremely low and those wrong answers happen to be all binned in the lowest possible confidence interval. This will provide misleading ECE result indicating that the model is ``perfectly'' calibrated. The second column of \texttt{gpt-3.5-turbo} in Figure \ref{fig:ece_plot} showcase such scenario.




\subsubsection{Tool-Use Collection}
\label{sec:appendix_tool_collection}
\begin{figure*}[h]
    \centering
    \includegraphics[width=\linewidth]{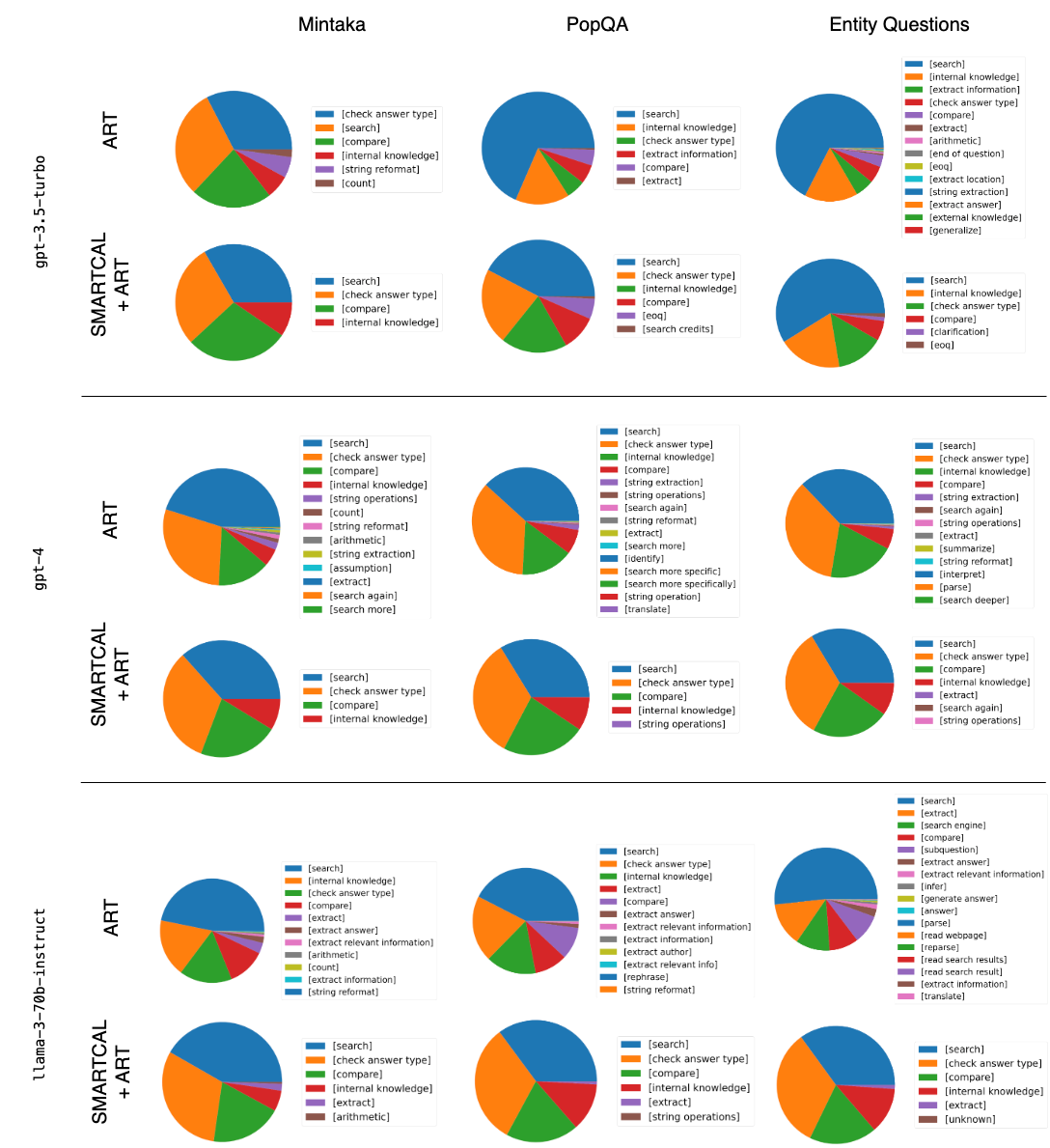}
    \caption{Tool-use comparison between ART and \textit{SMARTCAL}.}
    \label{fig:too_use_uncali}
\end{figure*}

We collect the tool usage distribution in both ART and \textit{SMARTCAL} for different models and demonstrate the results in Figure \ref{fig:too_use_uncali}.  There is a clear divergence in tool usage between ART and \textit{SMARTCAL}, where ART tends to include more tools that are unnecessary (such as ``\texttt{[string operations]}'' or ``\texttt{[code generate]}'') to augment its reasoning. The incorrect usage of tools often results in the introduction of redundant information in the context, which consequently degrades QA performance. 

\subsubsection{Calibration Curve Plot}
\label{sec:appendix_curve}
\begin{figure*}[h]
    \centering
    \includegraphics[width=\linewidth]{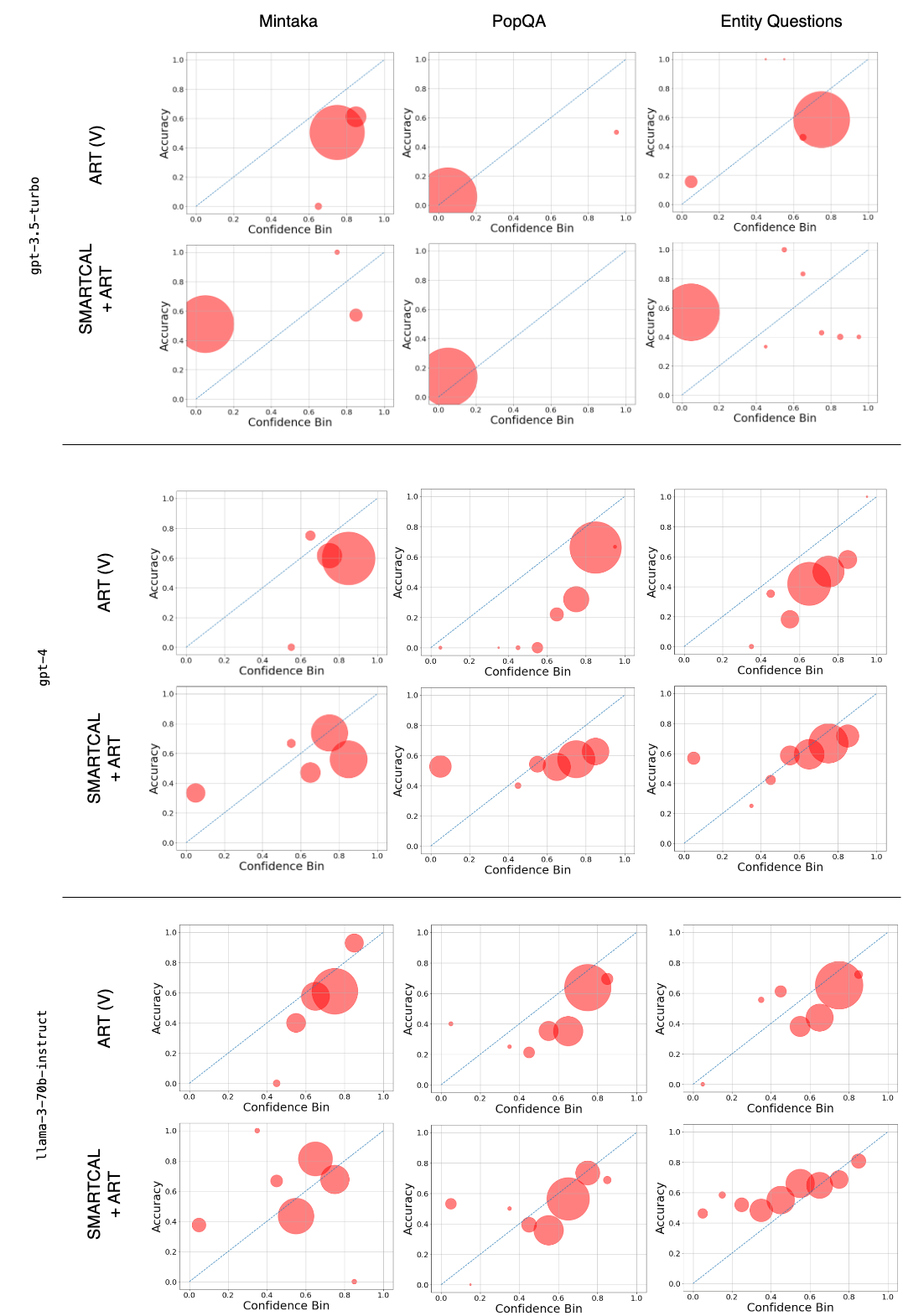}
    \caption{ECE plot comparison between ART (V) and \textit{SMARTCAL}.}
    \label{fig:ece_plot}
\end{figure*}

We also plot the ECE results for our framework on two approaches in Figure \ref{fig:ece_plot}. We select calibration results from ART (V) and compare them with ART augmented with \textit{SMARTCAL}. We segment the model stated confidence into 10 bins and calculate their QA accuracy with respect to each bin. We can see from the plot that under most cases, \textit{SMARTCAL} has a more sparse and aligned distribution along the reliance curve, i.e. the model stated confidence deviates less from the actual answer accuracy. 

\subsection{Prompts}
\label{sec:appendix_prompts}
In this section, we list the prompts that constitute the three major components in \textit{SMARTCAL} described in Section \ref{sec:framework}. We also provide ART (V) and DSP (V) prompts where we incorporate a verbalized calibration method that elicits model confidence on step-wise tool usage. For SE module, we curate three prompts, including task familiarity SE (Table \ref{tab:fam_se}), task similarity SE (Table \ref{tab:sim_se}), and tool-use instruction SE (Table \ref{tab:ins_se}). In our experiments, we use all three prompts in ART. Given that DSP only incorporates the retriever as the tool to use, we only use the task familiarity prompt in DSP. Note here for confidence prior collection phase in CPC, the prompt is essentially similar to prompts in ART (V) and DSP (V). For AR module, we include the calibration prompt in Table \ref{tab:cali_ar}.

\begin{table*}
    \centering
    \begin{tabular}{p{\linewidth}}
        \toprule
        \toprule
        \multicolumn{1}{c}{\textbf{DSP (V)}} \\
        \midrule
        Write a search query that will help answer a complex question. Write N/A if the context contains the answer to the question. Also include a confidence socre about your query. \\
        Note: The confidence level indicates the degree of certainty you have about your reasoning and is represented as a percentage. For instance, if your confidence level is 80, it means you are 80 percent certain that your answer is correct and there is a 20 percent chance that it may be incorrect.\\
        ---\\
        
        Follow the following format.\\
        Context:\${sources that may contain relevant content}\\
        
        Question: \${the question to be answered}\\
        
        Rationale: Let's think step by step. Based on the context, we have learned the following. \${a short summary from the context that provides useful clues} \\
        
        Search Query: \${a simple question for seeking the missing information}
        Confidence score: \${a score from 0 to 100} \\
        
        ---\\
        
        Context: \%s \\
        Question: \%s \\
        Rationale: Let's think step by step. Based on the context, we have learned the following.\\
        \bottomrule
        \bottomrule
    \end{tabular}
    \caption{Prompts in DSP (V) that incorporates verbalized confidence elicitation when using tools.}
    \label{tab:art_prompt}
\end{table*}

\begin{table*}
    \centering
    \begin{tabular}{p{\linewidth}}
        \toprule
        \toprule
        \multicolumn{1}{c}{\textbf{ART (V)}} \\
        \midrule
        In these examples, you are given a task description and an input. \\
        Break the input down into subtasks in order to solve the task. You can use affordances like string operations, search engines, arithmetic functions, or code generation. \\
        Be sure to use "[]" to specify affordances in subtasks. \\
        Also, use a separate '[]' to provide a score from 0 to 100 after each affordance to indicate your confidence level using this affordance. \\
        If you are confident that your internal knowledge is more reliable than external tools, use your own knowledge. \\
        When solving the task, avoid using affordances with low confidence level in the demonstrations below, because it often indicates a higher chance of making mistakes. If you still want to use them, make sure to assign a low confidence score. \\
    
        Note: The confidence level indicates the degree of certainty you have about your reasoning and is represented as a percentage. 
        For instance, if your confidence level is 80, it means you are 80 percent certain that your answer is correct and there is a 20 percent chance that it may be incorrect. \\
    
        ---- \\
        Selected Similar tasks: \%s \\
        ---- \\
        Description: \%s \\
        Input: \%s \\
        \bottomrule
        \bottomrule
    \end{tabular}
    \caption{Prompts in ART (V) that incorporates verbalized confidence elicitation when using tools.}
    \label{tab:verb_prompt}
\end{table*}

\begin{table*}
    \centering
    \begin{tabular}{p{\linewidth}}
        \toprule
        \toprule
        \multicolumn{1}{c}{\textbf{\textit{SMARTCAL} Task Familiarity SE}} \\
        \midrule
        Given a complex question to answer, determine whether using tools is necessary to answer it. If you determine that tools are unnecessary, you should include the suggestion to use "[Internal Knowledge]" only and downweight your confidence in using other tools. Otherwise you should provide a brief explanation on why tools are needed.\\
        ***\\
        Follow the following format:\\
        Task question: \${a complex question to answer}
        Familiarity verdict: \${Your verdict on whether to use tools. Often along with a brief explanation}
        ***\\
        
        Task question: \%s\\
        Familiarity verdict: \\
        \bottomrule
        \bottomrule
    \end{tabular}
    \caption{Task familiarity in the SE module of \textit{SMARTCAL}.}
    \label{tab:fam_se}
\end{table*}

\begin{table*}
    \centering
    \begin{tabular}{p{\linewidth}}
        \toprule
        \toprule
        \multicolumn{1}{c}{\textbf{\textit{SMARTCAL} Task Similarity SE}} \\
        \midrule
        You are given a question and several demos on using tools. Extract the name of the tools in the demos that you think are useful to answer the question. Don't select all tools, only include tools that you think are most helpful. Keep in mind to keep the tool list short. Note that tools are often expressed with their names in square brackets "[]". \\

        *** \\
        Follow the following format: \\
        Demo examples: \${few shot examples showing how to use different tools} \\
        Task question: \${a complex question to answer} \\ 
        Useful tools: \${a short list that keeps the minimal tools that helps answer the question. Remember to include a square bracket "[]" to any referred tool} \\
        *** \\
        
        Demo examples: \%s \\
        Task question: \%s \\
        Useful tools: \\
        \bottomrule
        \bottomrule
    \end{tabular}
    \caption{Task similarity in the SE module of \textit{SMARTCAL}.}
    \label{tab:sim_se}
\end{table*}

\begin{table*}
    \centering
    \begin{tabular}{p{\linewidth}}
        \toprule
        \toprule
        \multicolumn{1}{c}{\textbf{\textit{SMARTCAL} Tool-use Instruction SE}} \\
        \midrule
        Given the evaluation results on task similarity and familiarity, compile them into a detailed instruction that the agent can follow so that it can use tools more effectively. Make sure your instruction is based on the evaluation results and it should contain the following points: \\
        * Tell the agent whether or not it needs a tool \\
        * If no tool is needed, make sure to include [Internal Knowledge] in your reasoning \\
        * If needs a tool, always tell the exact name from the tool list in task similarity evaluation. Begin the instruction with "You should use..." \\
        * Include a square bracket "[]" for each tool that you tell the agent \\
        * Tell the agent not to use the tools not selected from the json file below \\
        * Provide the final instruction only, do not provide the previous evaluation results \\

        Below is a json file that describe the function of each tool \\
        ```json \\
        \%s \\
        ``` \\
        
        *** \\
        Follow the following structure by filling out the missing blocks with description: \\
        Evaluation results on task similarity: \${agent assessment on which tools are useful, often in a list expression} \\
        Evaluation results on task familiarity: \${agent assessment on tool confidence and verdict on whether to use its own knowledge} \\
        Instruction: Make sure you follow the following instructions before you move on. \${your verdict on whether to use own knowledge} You should use \${Tools from the similarity result} DO NOT use \${all tools not selected in similarity result but appeared in json file}. Keep using the right tools until you reach a final answer that is reliable. \\
        *** \\
        
        Evaluation results on task similarity: \%s \\
        Evaluation results on task familiarity: \%s\\
        Instruction:\\
        \bottomrule
        \bottomrule
    \end{tabular}
    \caption{Tool-use instruction in the SE module of \textit{SMARTCAL}.}
    \label{tab:ins_se}
\end{table*}



\begin{table*}
    \centering
    \begin{tabular}{p{\linewidth}}
        \toprule
        \toprule
        \multicolumn{1}{c}{\textbf{\textit{SMARTCAL} Calibration in AR}} \\
        \midrule
        You are given a resaoning process with confidence scores within each step in the square bracket "[]". \\
        Your job is to refer to the accuracy confidence table below and edit the confidence scores in the reasoning. \\
        Instructions: \\
        First identify the confidence range and find the corresponding accuracy in the table. If accuracy is lower than confidence, you should decrease the score. If accuracy is higher than confidence, you should increase the score. Finally, replace the original confidence score with your newly edited score. Your answer should keep the exact same structure of reasoning text and the input question, no extra explanation is needed. \\
        ----\\
        Below is the accuracy-confidence table:\\
        confidence level: \%s\\
        true accuracy: \%s\\
        ----\\
        Reasoning text to edit: \%s\\
        Your edited reasoning text:\\
        \bottomrule
        \bottomrule
    \end{tabular}
    \caption{Calibration prompt in AR module that enables collaboration between agents and confidence prior to recalibrate on tool-use.}
    \label{tab:cali_ar}
\end{table*}

\end{document}